\documentclass{article}

    \PassOptionsToPackage{numbers}{natbib}

\usepackage{fancybox}
\usepackage{longtable}
\usepackage{tabularx}
\usepackage{graphicx}
\usepackage{listings}
\usepackage{subcaption}

\usepackage[final]{neurips_2023}

\usepackage[utf8]{inputenc} 
\usepackage[T1]{fontenc}    
\usepackage{hyperref}       
\usepackage{url}            
\usepackage{booktabs}       
\usepackage{amsfonts}       
\usepackage{nicefrac}       
\usepackage{microtype}      
\usepackage{xcolor}         
\usepackage{graphicx}

\title{Function-constrained Program Synthesis}

\author{%
Patrick Hajali \quad Ignas Budvytis \\
University of Cambridge \\
\texttt{\{pah76,ib255\}@cam.ac.uk}
}

\begin{document}

\maketitle
\begin{abstract}
This work introduces: (1) a technique that allows pre-trained large language models (LLMs) to leverage user-provided code when solving programming tasks and (2) a method to iteratively generate modular sub-functions that can aid future code generation attempts when the initial code generated by the LLM is inadequate. Generating computer programs in general-purpose programming languages like Python poses a challenge for LLMs when restricted to using code provided in the prompt. A naive approach is to present a chat-based LLM (e.g. GPT-4) with relevant code snippets and prompt the model to synthesize the target algorithm using the provided code. Alternatively, code-specific LLMs (e.g. GitHub Copilot, CodeLlama2) can generate code completions in real-time by drawing on all code available in a development environment. However, restricting code-specific LLMs to use only in-context code is not straightforward, as the model is not explicitly instructed to use the user-generated code and users cannot highlight precisely which snippets of code the model should incorporate into its context for subsequent code-generations. Moreover, chat and code LLMs lack effective recovery methods, forcing users to iteratively re-prompt the model with modified prompts until a sufficient solution is reached.

Our method differs from traditional LLM-powered code-generation by constraining code-generation to an explicit function set and enabling recovery from failed attempts through automatically generated sub-functions. When the LLM cannot produce working code, we generate modular sub-functions to aid subsequent attempts at generating functional code. A by-product of our method is a library of reusable sub-functions that can solve related tasks, imitating a software team where efficiency scales with experience. 

We also introduce a new “half-shot” evaluation paradigm that provides tighter estimates of LLMs' coding abilities compared to traditional zero-shot evaluation. Our proposed evaluation method encourages models to output solutions in a structured format, decreasing syntax errors that can be mistaken for poor coding ability. 
\end{abstract}

\section{Introduction}
\begin{figure}[!t]
    \centering
    \begin{minipage}{0.43\textwidth}
        \centering
        \includegraphics[width=\linewidth]{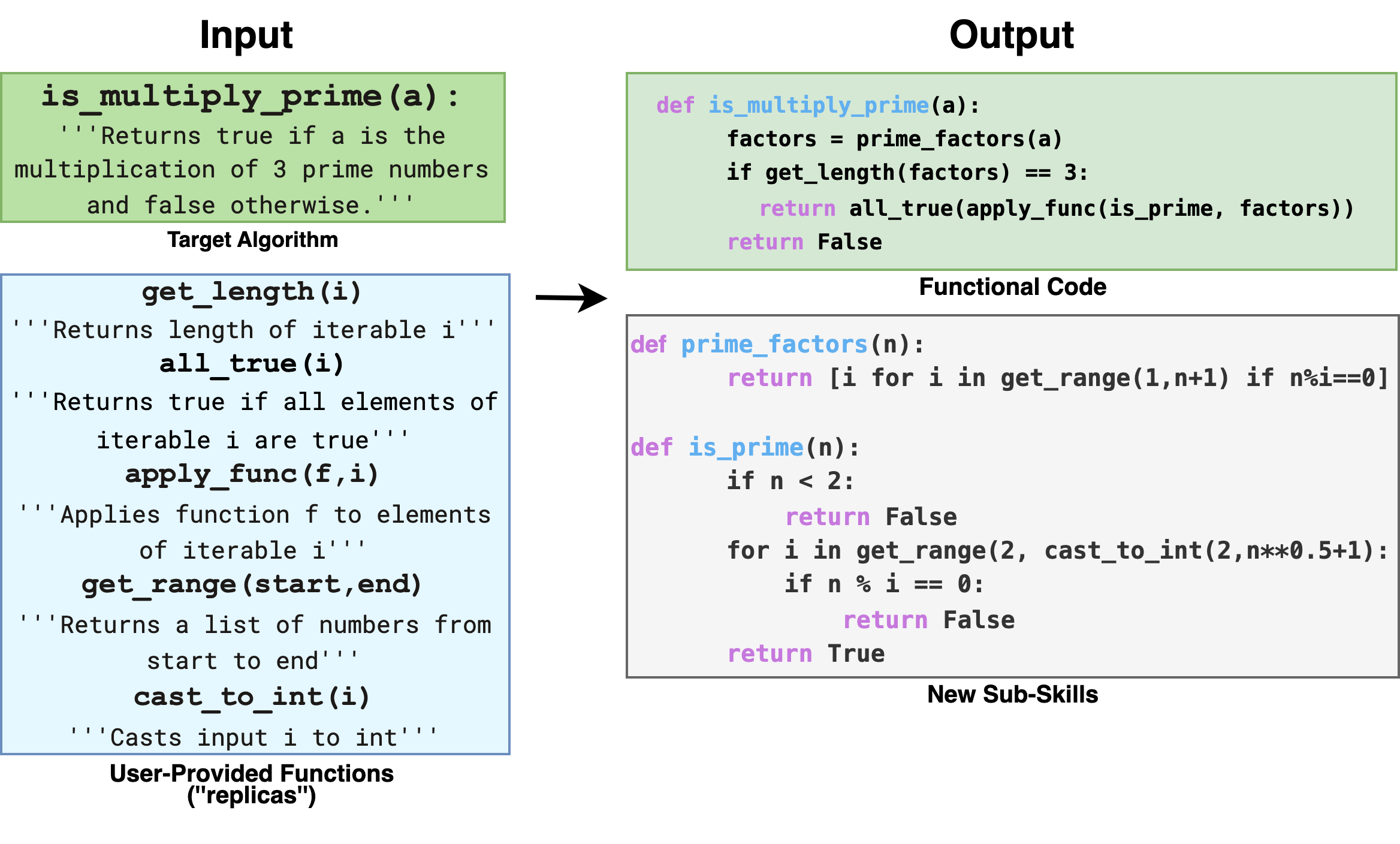}
        \subcaption{\label{fig:io}}
        
    \end{minipage}
    \hfill
    \begin{minipage}{0.56\textwidth}
        \centering
        \includegraphics[width=\linewidth]{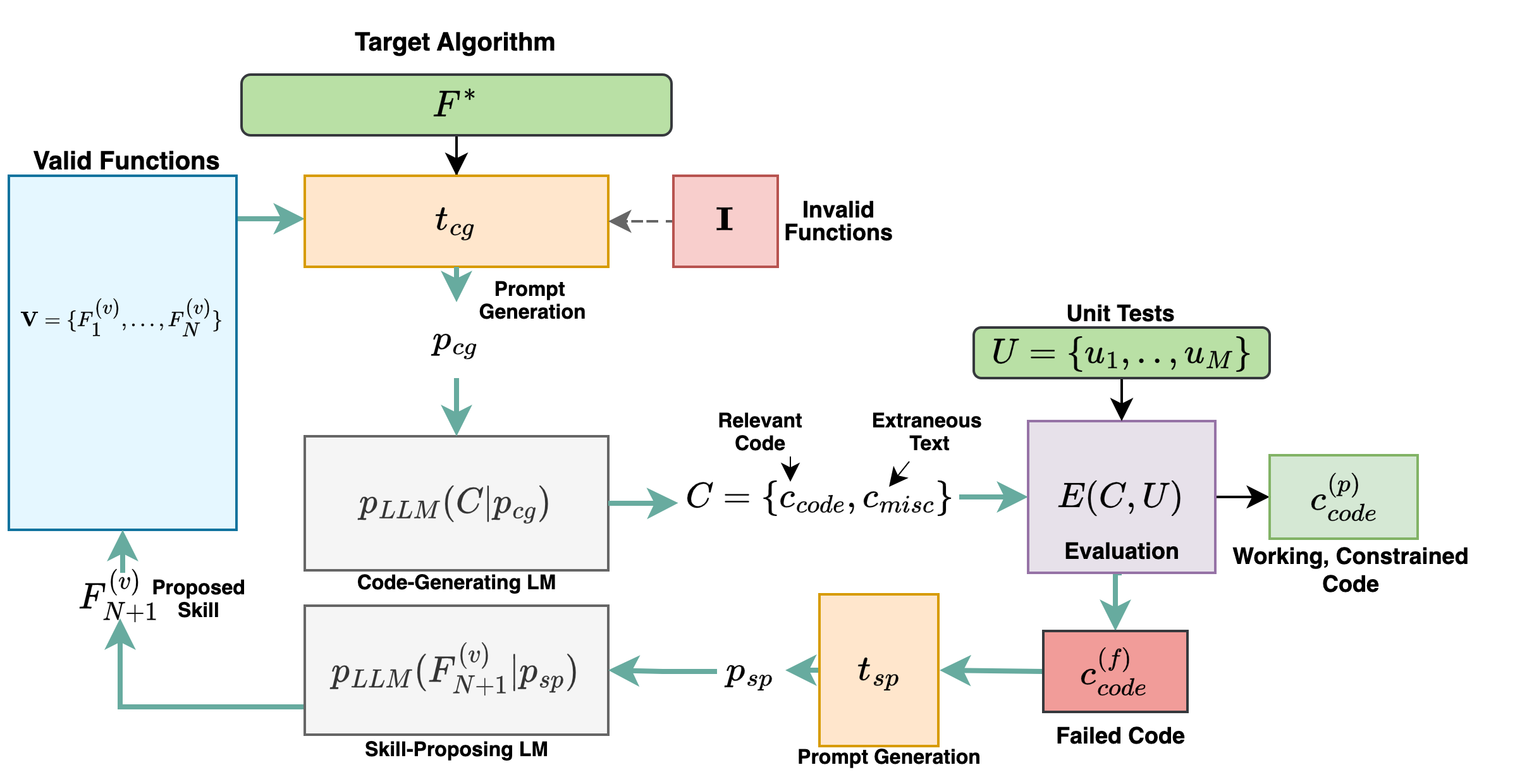}
        \subcaption{\label{fig:method}}
    \end{minipage}
    \caption{(a) Example input/output of our system on a question from HumanEval [\citenum{chen2021evaluating}]. Custom replicas, handwritten versions of common Python functions, are used to implement the target algorithm. Sub-functions are produced as a by-product. (b) Method: we constrain a LLM to implement a target algorithm $F^*$ using user-provided functions $V$ and basic Python operations. When the model fails, we generate modular sub-functions $F^{(v)}_{N+1}$ to add to the function set, iteratively building a library until the constrained model can generate working code. }
\end{figure}

Given code-snippets, in the form of functions, and a high-level algorithm description (in natural language), we aim to generate a program, in a general-purpose language (Python), implementing the algorithm using the given functions and basic language operations, as displayed in Figure \ref{fig:method}a. Doing so can enable automated integration of proprietary algorithms into existing infrastructure and can aid in bootstrapping development of new algorithms that build on existing code.

 Recent advances in large language models (LLMs) show promising capabilities for code comprehension, correction, and generation [\citenum{li2023explaining, bubeck2023sparks}] but are still limited in their ability to implement complex and compositional programs [\citenum{dziri2023faith}]. A reason for this limitation is the LLMs' inability to iteratively implement and reuse functions in realtime when generating code. To make use of unseen (during training) functions without retraining a LLM, the functions must be provided within the model's immediate context. Encouragingly, LLMs have demonstrated capabilities in using ``tools'', or API-calls, provided in-context [\citenum{schick2023toolformer, wu2023visual, shen2023hugginggpt, liang2023taskmatrix}], which is analogous to generating calls to \textit{functions} provided in-context. Motivated by this, we introduce here, to the best of our knowledge, the first methodology that not only primes LLMs to utilize in-context functions but also generates reusable sub-functions during program synthesis.

Designing a system that generates functional code while constrained to using user-specified functions presents several challenges. \textit{The main challenge is enabling the system to respond appropriately when it is initially unable to generate the desired algorithm}. In particular, the system requires a robust failure handling mechanism that can break down complex programs into more tractable components. While simple techniques like resampling or feeding the error trace back into the model [\citenum{chen2023teaching}] as additional input may allow the system to eventually produce functional code, these approaches are inefficient and risk the system getting stuck on challenging tasks. \textit{Another challenge is ensuring the model can aptly use a user-provided function} rather than reverting to behaviorally similar functions with which it is highly exposed to in training. \textit{A third challenge remains in ensuring that the output is presented in a usable format without requiring human postprocessing.} Pre-trained LLMs are fine-tuned to produce outputs in specific, non-standard formats. If these formats are not anticipated, avoidable syntax errors can arise when evaluating the generated code [\citenum{chen2023chatgpt}].

Recent works have proposed various solutions aimed at addressing these challenges. Approaches leveraging classical program synthesis methods have introduced the concept of ``library learning" [\citenum{DreamCoder, Bowers_2023}], exploring deductive algorithms which extract reusable components from an existing corpus of programs. While these works produce reusable function libraries as a byproduct of program synthesis, they often rely on domain-specific languages (DSLs) rather than general-purpose ones and require a heavy amount of input-output examples to specify the desired functions. LLMs such as GPT-4 [\citenum{openai2023gpt4}] can generate code from more relaxed natural language specifications combined with a few input-output examples. More commonly, LLMs are often used as coding "assistants" that suggest and debug lines of code in an auto-regressive manner. These models (e.g. GitHub Copilot [\citenum{github2021copilot}]) follow a limited range of context and produce code snippets in real-time to aid programmers. A coding-assistant, however, is designed to run alongside a human-programmer, and cannot be prompted to solve a number of problems on its own. Additionally, this approach does not currently allow the user to explicitly define which functions the LLM can and cannot use, making it difficult to ensure the model uses the desired functions. In a more autonomous fashion, ``agents" (e.g., AutoGPT [\citenum{auto_gpt_2023}], BabyAGI [\citenum{babyagi}], GPTEngineer [\citenum{gptengineer}],  Aider [\citenum{gauthier2023aider}]) can be initialized within a code-base and prompted to autonomously edit or generate code given the current context. Current agents, however, are not hard-coded with an effective way to handle cases when the model fails to produce working code and are prone to accumulation of compound hallucinations which can result in a loss of formatting structure and/or forgetting of the given functions. 

Our approach addresses the aforementioned challenges as follows. We extend library learning to general-purpose languages and embed algorithm decomposition as an iterative process, allowing a greater margin of error per attempt than methods which rely on an initial "planning" step to decompose the entire problem (e.g. SelfPlanning [\citenum{jiang2023selfplanning}]). To do this, we develop a prompting method to constrain a LLM to synthesise programs using only code (i.e., a set of functions) provided in-context, by the user. This allows the user to explicitly express which functions can and cannot be used in the generated code. If the constrained model fails to generate working code, we query an LLM to produce a new ``sub-function'' using only code provided in-context. The rationale is that the constrained model fails to generate working code because it lacks necessary functions in its context. We then add this sub-function to the set of valid functions to help the constrained model in future generation attempts - resulting in reusable functions as a byproduct - and iterate the process. A "half-shot" prompting and evaluation approach is introduced and used in all code-generation steps to ensure synthesize code is executable. We demonstrate the efficacy of our approach: achieving 73.1\% accuracy on HumanEval [\citenum{chen2021evaluating}] applying our method to GPT-3.5 [\citenum{brown2020language}] when constrained to an initial set of 21 hand-written functions designed to replicate behavior of common Python Standard Library (PSL) [\citenum{PythonStandardLibrary}] functions. We also generate solutions for 17.4\% questions in APPS [\citenum{HendrycksAPPS}] which were initially solved by the baseline GPT-3.5 but not solved by the constrained model even after 3 attempts. 

\section{Method} \label{seq:method}
In Subsection \ref{sec:overview}, we outline three key steps of our method: (1) constraining a LLM to code provided in-context via prompting, (2) generating sub-functions to aid the next iteration of code-generation, and (3) incorporating sub-functions into LLM-generated code. In Subsection \ref{sec:halfshot}, we introduce the "half-shot'' evaluation method. Examples for all prompts described are provided in supplementary material.

\subsection{Key Steps} \label{sec:overview}
\textbf{Constraining a LLM.} We begin by constraining a LLM to a set of user-provided functions $V$ and querying the constrained model to generate code. We model constrained code-generation as sampling code $C$ from an LLM (formulated as a probabilistic function and what to expect upon return) conditioned on prompt $p_{cg}$, which is generated by ``prompt-generator'' (or prompt-template) $t_{cg}$ (i.e. $C \sim p_{LLM}(C|p_{cg})$ where $p_{cg} = t_{cg}(F^*, V, I, p_f)$). Here, $t_{cg}$ takes as its inputs the target algorithm $F^*$, formatting-instruction $p_f$, the set of valid functions $V$, and the set of invalid functions $I$ as an input. The prompt generator $t_{cg}$ is designed to maximize the likelihood that the constrained model will generate working code which (1) calls the user-provided functions and (2) does not call any functions outside of the valid set $V$. To encourage correct usage of a user-provided function, the prompt must include, at minimum, instructions for calling each function and input/output behaviour (since the model needs to know how to call the function), although providing a description of the function and/or full source code may be beneficial. In $p_{cg}$, we specify that \textit{only} functions in $V$ may be called in generated code. To further maximize the second objective, explicitly listing invalid functions $I$ is useful when restricting use of functions highly exposed in LLM training (e.g. \texttt{len()}, \texttt{all()}; see Figure \ref{fig:io}). In summary, the prompt $p_{cg}$ is used to constrain the model and has the following high-level structure: a description of the target algorithm (e.g. a docstring), the set of valid user-provided functions (function names must be specified, but full code can also be provided), instructions to generate code utilizing only the valid user-provided functions (a set of invalid functions to explicitly restrict may also be specified), and formatting guidelines for the output code.

\textbf{Generating Sub-functions.} Code $C$ sampled from the constrained LLM is evaluated for functional-correctness. Evaluation is formulated as a function $E(C,U)$ which (1) parses relevant code $c_{code}$ in $C$ (removing any extraneous tokens, $c_{misc}$) and (2) evaluates $c_{code}$ against unit-tests $U = \{u_1, ..., u_M\}$ (see Section \ref{sec:halfshot} for more details). The code-evaluation block tags the code with a pass $(p)$ or fail $(f)$ and returns the tagged code as its output. When a constrained language model generates dysfunctional code (i.e. $c_{code}$ fails a single unit-test) for a given task $F^*$, our approach requires generating a new sub-skill which, when added to the set of valid functions, $V = \{F^{(v)}_1, ..., F^{(v)}_N\}$, is intended to increase the probability of sampling working code from the constrained language model. We define sub-skills (i.e. sub-functions) as modular Python functions that may be useful in implementing the desired program $F^*$. A new sub-function, when added to the collection of valid functions in the prompt, aims to enable the constrained model to generate working code in a subsequent attempt to implement the target algorithm. Sub-functions are proposed and implemented by sampling from a LLM conditioned on prompt $p_{sp}$: $F \sim p_{{LLM}}(F^{v}_{N+1}|p_{sp})$ where $p_{sp} = t_{sp}(c^{(f)}_{code}, F^*, V, I)$. Prompt $p_{sp}$ follows the same high-level structure as prompt $p_{cg}$ but instructs the model to generate a useful sub-function rather than code implementing the target algorithm. At minimum, the task description, $F^*$, valid functions $V$, and non-working code, $c^{(f)}_{code}$ are provided in prompt $p_{sp}$ for context. We also consider the case where working, \textit{non-constrained} (i.e. code which calls functions from $I$) code, $f^*_{soln}$, may be provided in-context in $p_{sp}$, serving as a reference to help predict a missing sub-function. In our experiments, $f^*_{soln}$ is provided by the dataset or produced by an unconstrained language model. 

\textbf{Integrating Sub-Functions into LLM-Generated Code.} To integrate a new sub-skill $F^{(v)}_{N+1}$ into subsequent code-generations we can simply add it to the set of valid functions $V$ (i.e. $V \leftarrow \{F^{(v)}_{N+1}\}\cup V$; $V$ is dynamically updated with the proposed sub-skill). This alone, however, is not sufficient. Because $F^{(v)}_{N+1}$ was implemented to aid specifically in solving algorithm $F^*$, it is likely that $F^{(v)}_{N+1}$ is more useful than other functions in the context of solving $F^*$, and should be distinguished within prompt $p_{cg}$, among $F^{(v)}\in V$. To do this, we introduce $V^*\subset V$ where $V^*$ contains all $F^{(v)}\in V$ which have been specifically proposed to aid implementing the target algorithm $F^*$. We then formulate the new prompt with respect to this distinction: $p_{cg} = \hat{t}_{cg}(F^*, V^*, V, I, p_f)$. 
The updated prompt $p_{cg}$ is used to condition the language model, from which a new solution to $F^*$ is sampled. In our experiments, we only provide a single sub-function at-a-time ($|V^*|=1$), although we investigate the effects of providing multiple sub-functions ($|V^*|>1$) in an ablation included in supplementary material. 

\subsection{Half-Shot Evaluation} \label{sec:halfshot}

Typically, datasets like HumanEval [\citenum{chen2021evaluating}] and APPS [\citenum{HendrycksAPPS}] provide a prompt describing the code to implement, which is fed directly to the code-generating model as the \textit{only input} when evaluating for zero-shot performance. In this context ``zero-shot'' means that the only input to the model is the prompt provided by the dataset. The model's output is evaluated for functional correctness, generally through unit-testing. Pass@k accuracy [\citenum{passk}] is reported. This evaluation methodology is problematic when models have inconsistent output formatting. Specifically, top-performing LLMs (e.g., GPT-4 [\citenum{openai2023gpt4}], Claude-2 [\citenum{anthropic2023claude2}]) are often finetuned to behave interactively, generating conversational responses rather than completions. This causes variability in how outputs are formatted by default - some models tend to wrap code in markdown formatting like ```triple backticks''' or may include explanatory text before or after the code. These extraneous characters result in syntax errors when the output is evaluated, penalizing the model's score regardless of the functionality of the code provided [\citenum{chen2023chatgpt}]. 

In Table \ref{fig:eval_motiv_results}, pass@1 accuracy on HumanEval [\citenum{chen2021evaluating}] with temperature set to 0 is reported using three prompt templates. Both GPT-4 and GPT-3.5 improve markedly when stricter formatting instructions are used, indicating \textit{many zero-shot failures may be due to inaccurate formatting rather than coding inability.} For reference, the top performing model, Reflexion [\citenum{shinn2023reflexion}], achieves 91\% accuracy on HumanEval. Other models, such as Parsel [\citenum{zelikman2022parsel}] (which achieves 85.1\% accuracy), provide no additional performance beyond what a formatting-string and parser can provide. 

\begin{table}[t]
    \centering
    \small
    \caption{Accuracy (\textit{pass@1}) on HumanEval [\citenum{chen2021evaluating}] with varying levels of formatting instruction provided in the prompt. Zero-shot prompt only includes the prompt provided by the dataset prompt (i.e.  a docstring). The ``basic'' prompt adds "Complete the code." to the system prompt. The formatting-prompt/parser provides specific format instructions and parses outputs expecting that format. Both models perform drastically better when given stricter formatting instructions.
    }
    \label{fig:eval_motiv_results}

    \footnotesize
    \begin{tabular}{cccc}
    \toprule
    & \textbf{Zero-shot} & \textbf{Basic} & \textbf{Formatting-string with parser (half-shot)} \\
    \midrule
    \textbf{GPT-4} & 67.1\% & 73.7\% & 85.4\% \\ 
    \textbf{GPT-3.5} & 34.1\% & 70.0\%  & 71.3\% \\ 
    \bottomrule
    \end{tabular}


\end{table}


To get a better, unbiased estimate of models' baseline coding-performance,  we introduce a modified evaluation method focused on establishing upper bound estimates for a models coding-ability. Our evaluation approach has three-steps: (1) Add a formatting-string to the prompt given by the dataset, instructing the model to formulate relevant code in as easy-to-parse format; (2) Parse the model output to extract the relevant code. The parser should strip non-code from model outputs before evaluation, minimizing syntax errors; (3) Evaluate the extracted code. In comparison to ``zero-shot'' evaluation, ``half-shot'' evaluation simply adds a formatting-string to the model input and passes the model's output through a parser. 
Although our evaluation method does not qualify as zero-shot, we maintain that no task-specific informational advantages are provided in-context in the evaluation step.

\section{Experiments and Results}
In this section, we first describe the datasets and sub-datasets used in our experiments. We then present the experiments leading to three key findings: (1) constraining a pre-trained LLM to unseen functions provided in-context degrades coding performance, (2) performance can be recovered by introducing appropriate sub-functions to the in-context function set, and (3) LLMs can automatically generate these useful sub-functions.

\textbf{Datasets.} Along with HumanEval (HE) [\citenum{chen2021evaluating}] and APPS [\citenum{HendrycksAPPS}], four sub-datasets are relevant in our experiments: $HE_{CF}$ contains 8 HumanEval questions which were answered correctly by the baseline GPT-3.5 model in a single attempt but failed when the model was constrained, even after three solutions were sampled at temperature ($\epsilon$) set to 0.5; $HE_{BFF}$ contains 19 HumanEval questions that neither the unconstrained baseline nor the constrained GPT-3.5 model could answer, even after three solutions were sampled with $\epsilon = 0.5$; \(HE_{BP,R}\) comprises 69 questions solved by the baseline unconstrained GPT-3.5 model (with $\epsilon = 0$) that feature a free-standing function call in their solution (this set helps formulate \(V_{rep}\) described below) $APPS_{BP}$ contains 1329 questions from the APPS dataset which the unconstrained GPT-3.5 model successfully solved in a single attempt with $\epsilon = 0$.


    


\begin{table}[]
    \caption{''Half-shot'' accuracy (\textit{pass@1}) on HumanEval [\citenum{chen2021evaluating}], $HE_{BP,R}$, and APPS [\citenum{HendrycksAPPS}] before and after constraining models' to only generate calls to functions ($V_{rep}$) provided in-context.}
    \centering
    \footnotesize

    \begin{tabular}{lcccc}
    \toprule
    & \textbf{HumanEval} & \textbf{$\mathbf{HE_{BP,R}}$} &  \textbf{$\mathbf{APPS_{BP}}$} \\
    \midrule
    \textbf{GPT-4} & 85.4\% & 94.2\%  &  -\\ 
    \textbf{GPT-4 Constrained} & 73.9\% & 78.2\%  &  -\\ 

    \midrule
    
    \textbf{GPT-3.5} & 71.3\% & 100\%   &   100\% \\ 
        \textbf{GPT-3.5 Constrained} & 65.8\% & 75.3\%  &  10.4\% \\ 
    \bottomrule
    \end{tabular}

    
    \label{fig:const_results_table}
\end{table}
\textbf{Constraining a LLM.} We constrain both GPT-3.5 and GPT-4 to a set of functions denoted by \( V_{rep}\) which consists of 21 hand-written functions (e.g. \texttt{get\_length}, \texttt{all\_true} from Figure \ref{fig:io}) designed to emulate the behavior of specific functions in the PSL, but under different names (e.g. \texttt{len()} becomes \texttt{get\_length()}). This set of functions are referred to as "replicas," with each corresponding to an "origin function" in the PSL (details on all replicas is provided in supplementary material). The replicas are derived from all origin functions found in solutions to questions in \( HE_{BP,R} \). Thus, constraining a model to using only replicas ensures no necessary functions are withheld when solving questions specifically in \( HE_{BP,R} \). We constrain the models through prompting where prompts are constructed per-question from prompt generator $t_{cg}$ as described in Section \ref{seq:method}. We design $t_{cg}$ (details provided in supplementary material) to maximize the Utilization Rate (UR; percent of solutions in which a function provided in-context is called) and minimize the Non-Compliance Rate (NCR; percent of solutions in which a function \textit{not} provided in-context is called). Using the optimal $t_{cg}$, GPT-4 achieves a UR of 92.1\% and NCR of 42.7\% on HE [\citenum{chen2021evaluating}]. \textit{The model utilizes the provided functions often, although it frequently invokes invalid functions as well.} This is to be expected, since the original functions we replicated are among the most commonly used Python functions.

Having found an effective way to constrain GPT-3.5/4, we asses the functionality of the generated code. Table \ref{fig:const_results_table} provides the ``half-shot" pass@1 accuracy of GPT-3.5 and GPT-4 with and without a constraint on various datasets. Importantly, we do not mark code incorrect specifically for using invalid functions. This allows us to independently evaluate change in the model's coding performance separately from its capacity to follow the constraint. On $HE_{BP,R}$, the accuracy of GPT-3.5 declines by 24.7\% when constrained. This suggests that the constrained models' decrease in performance is not caused by restricting access to necessary functions, but rather from forcing the model to use functions it has not seen in training. We analyzed all questions in $HE_{BP,R}$ which were passed by unconstrained GPT-3.5 but failed by constrained GPT-3.5 and found that none of the newly failed questions were a result of syntax errors (details provided in supplementary material). The decline in accuracy for the entire HumanEval dataset, -11.5\% for GPT-4, is less marked than in \(HE_{BP,R}\) because questions present in HumanEval but absent in \(HE_{BP,R}\) are either already failed by the unconstrained baseline or solvable without the necessity of a replica. We attribute the 90\% performance drop on the $APPS_{BP}$ dataset to the higher complexity of APPS questions compared to HumanEval. It is likely that the unconstrained model's baseline performance is inherently unstable for these questions.  In summary, we find that GPT-3.5 and 4 perform worse on coding-tasks when they are instructed to generate code which can only contain calls to unseen functions provided in-context.



\begin{table}[t]
\caption{GPT-3.5 constrained on replicas pass@1 accuracy after adding sub-skills to the set of usable functions provided in-context. We compare the efficacy of sub-skills generated by GPT-4, GPT-3.5 and a human expert, with and without (unconstrained) ground truth code $f^*_{soln}$ provided for reference. Gain in parantheses refers to improvement above the baseline model (unconstrained GPT-3.5). Note that the baseline model failed all questions in $HE_{CF}, HE_{BF}$.} 
\centering
\footnotesize
    \begin{tabular}{lccccc}
        \toprule
        $f^*_{soln}$  & Skill-Proposing LLM & \textbf{HE} & \textbf{$\mathbf{HE_{CF}}$} & \textbf{$\mathbf{HE_{BFF}}$} & \textbf{$\mathbf{APPS_{BP}}$} \\
         \midrule
        \textbf{Yes}& \textbf{Human Expert} & - & 100\% & - & - \\
        & \textbf{GPT-4} & - & 62.5\% & 31.6\% & - \\
        & \textbf{GPT-3.5} & - & 37.5\% & 15.8\% & -  \\
        
        \midrule
        \textbf{No} & \textbf{GPT-4} & 73.1\%(+7.3) & 50\% & 42.1\% & - \\
        & \textbf{GPT-3.5} & 72.5\% (+6.7) & 62.5\% & 10.5\% & 17.6\% (+7.2) \\
       
        \bottomrule
    \end{tabular}
    \label{fig:skill_results_table}
\end{table}

\textbf{Providing Sub-Functions to the Constrained Models to Recover Performance.} Here, we demonstrate that introducing sub-functions, as detailed in Section \ref{seq:method}, can ameliorate some of the performance loss experienced after constraining the models. For each question in $HE_{CF}$ we propose and hand-write a useful Python function which, when added to the set of functions provided in-context, increases the likelihood of drawing functional code from the constrained code-generating model when sampling a new solution to the previously failed question. As shown in the bottom row of Table \ref{fig:skill_results_table}, \textit{we are able to recover lost performance on 100\% of questions in $HE_{CF}$ by providing hand-written sub-functions.} Note that the "human expert'' meticulously crafted (through trial and error) the ideal sub-functions tailored for the \(HE_{CF}\) dataset to serve as an example of how performance can  be fully restored given optimal sub-functions. Writing these functions by hand, however, is not scalable.

\textbf{Generating Sub-Functions Automatically.} Here, we demonstrate that sub-function generation can be scaled efficiently using LLMs. We employ GPT-3.5 and GPT-4 to independently propose and implement sub-functions intended help constrained models producing working code for questions in HumanEval [\citenum{chen2021evaluating}], $HE_{CF}$, $HE_{BFF}$ and $APPS_{BP}$. As in the previous experiment, each sub-skill is added to the set of functions provided in-context to GPT-3.5. Table \ref{fig:skill_results_table} presents the "half-shot" pass@1 accuracy observed when sub-functions are provided. We consider the case when the skill-generating model is and is not provided ground-truth, unconstrained code \(f^*_{soln}\). While LLM-generated sub-functions are not as effective as those generated by a human-expert (GPT-4's sub-functions result in +62.5\% on \(HE_{CF}\)), LLM-generated sub-functions address challenges previously impassible by the constrained model alone. Impressively, providing automatically-generated sub-functions resolves nearly half of the questions that even the unconstrained model initially faltered upon (+42.1\% on \(HE_{BFF}\)). The constrained GPT-3.5 model is also able to answer 73.1\% of questions right on the HumanEval [\citenum{chen2021evaluating}] dataset when GPT-4 produced sub-functions are provided in-context, surpassing the performance of the unconstrained GPT-3.5 model.

\section{Conclusion}

This report introduces a novel method for constrained program synthesis in a general purpose programming language using LLMs.  Our experiments demonstrate that constraining LLMs like GPT-3.5 and GPT-4 to unseen functions provided in-context significantly reduces their ability to generate working code, with performance dropping by 11-89\% across datasets. This highlights the limitations of LLMs when forced to use unfamiliar functions. To address this, we develop an iterative process to generate sub-functions that aid the model in solving tasks it previously failed on. Providing sub-functions as this additional context enables the constrained models to recover much of the lost performance. Both GPT-3.5 and GPT-4 are able to automatically generate effective sub-functions that improve constrained model performance, increasing constrained GPT-3.5's accuracy on questions previously failed after 3 attempts by 62.5\%. 

\bibliography{references}
\bibliographystyle{IEEEtran}

\newpage
\appendix 
\section{Supplementary Material}

We provide the following: 
\begin{itemize}
    \item In Section \ref{prompts}, we present examples of prompts $p_{cg}$ and $p_{sp}$ used for code-generation and skill-proposals, respectively. We also provide examples of the prompts and parser used in motivating Half-Shot evaluation.
    \item In Section \ref{replicas}, we present a list of all the replicas and their hand-written code. 
    \item In Section \ref{exs}, we qualitatively compare outputs from GPT-3.5 before and after a constraint is applied. 
    \item In Section \ref{fails}, we present a detailed report of all questions in $HE_{BP,R}$ failed by GPT-3.5 when constrained to the replicas.
    \item In Section \ref{ablations}, we present details of an ablation on the size of $V^*$ briefly mentioned in the main paper. 
    \item In Section \ref{datasets}, we present a graphical overview of the sub-datasets used in our experiments. 
\end{itemize}

\subsection{Prompts} \label{prompts}

\begin{figure}[!h]

\shadowbox{%
  \begin{minipage}{\dimexpr\textwidth-2\fboxrule-2\fboxsep}
    \begin{center}\textbf{System Prompt} \end{center}  
    Follow the user instructions and provide an implementation of what you deem to be the most useful sub-function.
  \end{minipage}%
}

\vspace{7pt} 

\shadowbox{%
  \begin{minipage}{\dimexpr\textwidth-2\fboxrule-2\fboxsep}
    \begin{center}\textbf{User Prompt} \end{center} 
I have the following docstring:\\
\texttt{\{docstring\}}\\
\\
A correct completion to this function is: \\
\texttt{\{unconst\_comp\}} \\

I constrained a language model to generate a new completion using only custom Python functions that I provided.
I gave it access to the following functions: \\
\texttt{\{valid\_functions\}}\\

It generated the following incorrect completion: \\
\texttt{\{failed\_completion\}} \\

Note that none of the following functions are allowed: \\ 
\texttt{\{restricted\_functions\}} \\ 

What new function would be useful to provide to the “constrained” language model to help it produce a working completion?
Propose a completely new function. Only output code implementing the new function you propose. Only output executable code. Format your answer as follows: \\

\# BEGIN NEW-SUB FUNCTION
… 

  \end{minipage}%
}

\caption{\textbf{Example skill-proposal prompt for an OpenAI chat model (GPT4/3.5).} This illustrates querying for a single sub-skill function, though the prompt could be modified to request multiple skills. In our experiments, \texttt{valid\_functions} ($\mathbf{V}$) are the replicas (full code is provided) and \texttt{restricted\_functions} ($\mathbf{I}$) is a list of the names of all the associated origin functions. The \texttt{docstring} ($\mathbf{F^*}$) and \texttt{failed\_completion} ($c^{(f)}_{code}$) are included to provide context. Working code ($f^*_{soln}$) which does not adhere to the constraints is provided in this example.}
\label{prompt:proposal}

\end{figure}
\begin{figure}[!h]

\centering
 
\shadowbox{%
  \begin{minipage}{\dimexpr\textwidth-2\fboxrule-2\fboxsep}
    \begin{center}\textbf{System Prompt} \end{center}  
    You are an intelligent programmer. You must complete the python function given to you by the user using only the functions they give you. And you must follow the format they present when giving your answer!
  \end{minipage}%
}

\vspace{7pt} 

\shadowbox{%
  \begin{minipage}{\dimexpr\textwidth-2\fboxrule-2\fboxsep}
    \begin{center}\textbf{User Prompt} \end{center} 
You have access to the following Python functions:\\

\# VALID FUNCTIONS\\ 
\texttt{\{valid\_functions\}} \\ 

The following functions may be particularly useful:\\

\# RELEVANT FUNCTIONS\\ 
\texttt{\{relevant\_functions\}} \\ 

You must complete the python function I give you using ONLY the given functions. You CANNOT use any of the following invalid functions: \\

\# INVALID FUNCTIONS\\
\texttt{\{restriced\_functions\}} \\

You must write the completion in the following form:\\

\# FUNCTION HEADER\\
...\\
\# START OF COMPLETION\\
...\\

You may only write your response in code/comments. Do not be verbose. The function you are to complete is:\\

\texttt{\{docstring\}}

  \end{minipage}%
}

\caption{\textbf{Example code-generating prompt $p_{cg}$ for an OpenAI chat model (GPT4/3.5).} Constraints on valid and invalid functions are provided in the User Prompt.  Alternatively, constraints can be specified in the Sytem Prompt. In our experiments, \texttt{valid\_functions} ($\mathbf{V}$) are the replicas (full code is provided), \texttt{restricted\_functions} ($\mathbf{I}$) is a list of the names of all the associated origin functions, and \texttt{docstring} ($\mathbf{F^*})$ describes the target algorithm. Particularly \texttt{relevant\_functions} ($\mathbf{V^*}$) previously proposed in a function-generation step are distinguished. Using natural, commenting-style formatting instructions increases likelihood of correct formatting.}
\label{prompt:constraint}

\end{figure}
\begin{figure}[!h]
    \centering
    \includegraphics[width=0.8\linewidth]{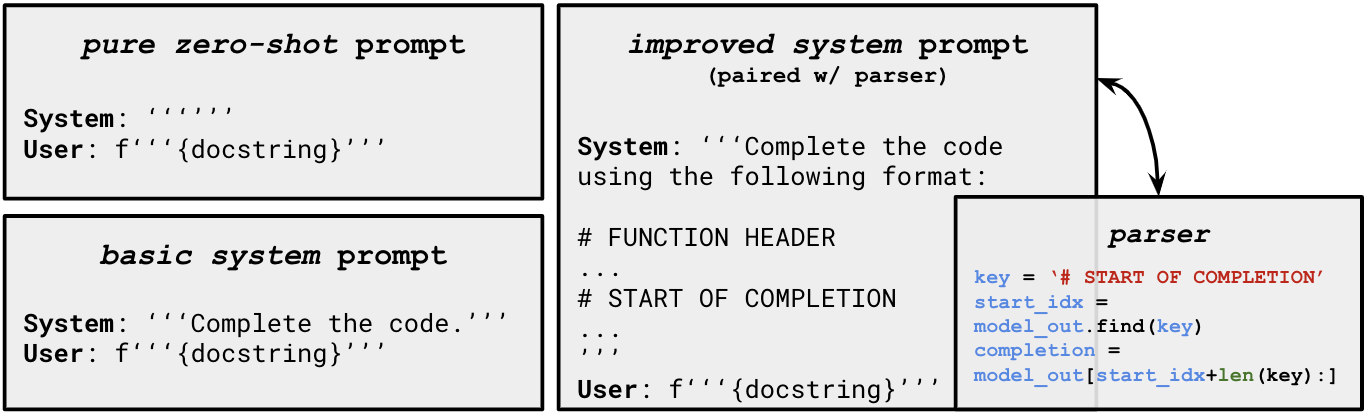}
    \caption{\textbf{Prompts used to Motivate Half-Shot Evaluation.} Basic zero-shot prompt (top left) only includes dataset prompt, i.e., \texttt{docstring}. Next level (bottom left) adds "Complete the code." to the system prompt. Final prompt (right) provides specific format instructions and parses outputs expecting that format.} 
    \label{fig:eval_motiv}
\end{figure}

\newpage 
\subsection{Replicas} \label{replicas}

In this section, we provide details on the specific functions used to constrain the models in our experiments. The models are restricted to a particular set of functions, denoted by \(V_{rep}\). A detailed overview of these functions can be found in Table \ref{tab:replicas}.

\lstset{
  language=Python,
  basicstyle=\ttfamily\tiny,
  breaklines=true,
  showstringspaces=false,
  aboveskip=0.2mm,
  belowskip=0mm,
  emph={is_prime(next_prime), is_prime(length), is_prime}, 
  emphstyle=\bfseries, 
}

\begin{longtable}{|l|l|}
\caption{Hand-written code used for the Replicas. The corresponding function from the Python Standard Library is given in the left column.} \label{tab:replicas} \\
\hline
\textbf{Original function} & \textbf{Custom re-implementation (Replica)} \\
\hline
\endhead  

len() & \begin{lstlisting}
def get_length(iterable):
    count = 0
    for _ in iterable:
        count += 1
    return count
\end{lstlisting} \\
\hline
str() & \begin{lstlisting}
def cast_to_string(input):
    return str(input)
\end{lstlisting} \\
\hline
chr() & \begin{lstlisting}
def convert_to_char(input):
    return chr(input)
\end{lstlisting} \\
\hline
float() & \begin{lstlisting}
def cast_to_float(input):
    return float(input)
\end{lstlisting} \\
\hline
int() & \begin{lstlisting}
def cast_to_int(input):
    return int(input)
\end{lstlisting} \\
\hline
list() & \begin{lstlisting}
def create_list(iterable=None):
    if iterable is None:
        return []
    lst = []
    for item in iterable: lst.append(item)
    return lst
\end{lstlisting} \\
\hline
set() & \begin{lstlisting}
def create_set(iterable=None):
    s = {}
    if iterable:
        for element in iterable: s[element] = None
    return s.keys()
\end{lstlisting} \\
\hline
isinstance() & \begin{lstlisting}
def check_if_instance(obj, class_or_tuple):
    if not isinstance(class_or_tuple, tuple):
        class_or_tuple = (class_or_tuple,)
    for cls in class_or_tuple:
        if type(obj) == cls or type(obj) in cls.__subclasses__():
            return True
    return False
\end{lstlisting} \\
\hline
sorted() & \begin{lstlisting}
def sort_list(iterable, key=None, reverse=False):
    lst = list(iterable)
    if key is None:
        compare = lambda a, b: a > b
    else:
        compare = lambda a, b: key(a) > key(b)
    for i in range(len(lst)):
        for j in range(len(lst) - 1):
            if compare(lst[j], lst[j + 1]):
                lst[j], lst[j + 1] = lst[j + 1], lst[j]
    if reverse:
        lst = lst[::-1]
    return lst
\end{lstlisting} \\
\hline
all() & \begin{lstlisting}
def check_if_all_true(iterable):
    for element in iterable:
        if not element:
            return False
    return True
\end{lstlisting} \\
\hline
min() & \begin{lstlisting}
def get_minimum(*args): 
    if len(args) == 1:
        args = args[0]  
    if not args:
        raise TypeError('expected at least 1 arguments, got 0')
    min_val = args[0]
    for arg in args:
        if arg < min_val:
            min_val = arg
    return min_val
\end{lstlisting} \\
\hline
max() & \begin{lstlisting}
def get_maximum(*args):
    ...
    return max_val
\end{lstlisting} \\
\hline
bin() & \begin{lstlisting}
def convert_to_binary(n):
    if n < 0:
        return '-' + convert_to_binary(-n)
    result = ''
    while n:
        result = ('1' if n & 1 else '0') + result
        n >>= 1
    return '0b' + result if result else '0b0'
\end{lstlisting} \\
\hline
sum() & \begin{lstlisting}
def compute_sum(iterable, start=0):
    total = start
    for item in iterable:
        total += item
    return total
\end{lstlisting} \\
\hline
round() & \begin{lstlisting}
def round_number(number, ndigits=None):
    if ndigits is None:
        return int(number + 0.5) if number >= 0 else int(number - 0.5)
    else:
        factor = 10.0 ** ndigits
        return int(number * factor + 0.5 if number >= 0 else number * factor - 0.5) / factor
\end{lstlisting} \\
\hline
math.ceil() & \begin{lstlisting}
def get_ceiling(number):
    integer_part = int(number)
    if number == integer_part:
        return integer_part
    if number > 0:
        integer_part += 1
    return integer_part
\end{lstlisting} \\
\hline
math.sqrt() & \begin{lstlisting}
def get_square_root(input, precision = 0.00001):
    guess = input / 2.0 
    while True:
        better_guess = (guess + input / guess) / 2.0
        if abs(guess - better_guess) < precision: 
            return better_guess
        guess = better_guess
\end{lstlisting} \\
\hline
ord() & \begin{lstlisting}
def get_unicode(char):
    if len(char) != 1:
        raise TypeError("Error." % len(char))
    return int.from_bytes(char.encode('utf-8'), byteorder='big')
\end{lstlisting} \\
\hline
map() & \begin{lstlisting}
def apply_func_to_iterable(function, iterable):
    result = []
    for item in iterable:
        result.append(function(item))
    return result
\end{lstlisting} \\
\hline
abs() & \begin{lstlisting}
def absolute_value(number):
    if number < 0:
        return -number
    else:
        return number
\end{lstlisting} \\
\hline
filter() & \begin{lstlisting}
def add_to_list_if_func_is_true(function, iterable):
    result = []
    for item in iterable: if function(item): result.append(item)
    return result
\end{lstlisting}\\
\hline
\end{longtable}

\(V_{rep}\) comprises 21 hand-written functions created to emulate common functions from the Python Standard Library. Hence, we call these functions ``replicas." They are modeled on functions utilized by the unconstrained, baseline model during its evaluation on the HumanEval dataset. All replicas adhere to the following criteria:

\begin{enumerate}
    \item A replica must maintain the same functionality as the original function it aims to imitate.
    \item A replica must bear a distinct name from the corresponding original function, although the name should reflect its purpose. For instance, the replica mimicking \texttt{len} is named \texttt{get\_length}.
\end{enumerate}

 In our experiments, we want to constrain the models to ``replicas" in specific to ensure that we are not depriving the models of any essential resources they need to solve the problems in our dataset.

\newpage
\subsection{Examples of Model Output: Constrained vs. Unconstrained} \label{exs}

\begin{figure}[!h]
\lstset{
    language=Python,  
    basicstyle=\ttfamily,
    keywordstyle=\color{blue},
    commentstyle=\color[rgb]{0.13,0.55,0.13},
    stringstyle=\color{red},
    showstringspaces=false,
    numberstyle=\tiny,
    breaklines=true,
}

\shadowbox{
    \begin{minipage}{\textwidth}

    \begin{center}\textbf{Docstring}\end{center} 
    \begin{lstlisting}
def cycpattern_check(a , b):
    '''
    You are given 2 words. You need to return True if the second word or any of its rotations is a substring in the first word 
    cycpattern_check("abcd","abd") => False  
    cycpattern_check("hello","ell") => True
    cycpattern_check("whassup","psus") => False
    '''
    \end{lstlisting}
    \end{minipage}
}

\vspace{1em}

\begin{minipage}{0.47\textwidth}
    \shadowbox{
        \begin{minipage}{\textwidth}
        \begin{center}\textbf{Unconstrained-Model Output} \end{center}
        \begin{lstlisting}
if len(a) < len(b):
    return False
if b in a:
    return True
for i in range(1, len(b)):
    if b[i:] + b[:i] in a:
        return True
return False
\end{lstlisting}
        
        \end{minipage}
    }
\end{minipage}
\hfill
\begin{minipage}{0.47\textwidth}
    \shadowbox{
        \begin{minipage}{\textwidth}
        \begin{center}\textbf{Constrained-Model Output} \end{center} 
\begin{lstlisting}
# Check if the length of b is greater than the length of a 
if get_length(b)>get_length(a):
    return False 
# Concatenate a with itself to check for rotations of b 
a = a + a
# Check if b is a substring of a or any of its rotations
if b in a:
    return True
return False
        \end{lstlisting}

        \end{minipage}
    }
\end{minipage}

\caption{\textbf{Comparing Outputs from Constrained and Unconstrained GPT-3.5.} This figure shows the responses of GPT-3.5, in both constrained and unconstrained conditions, to a particular question involving the detection of cyclic patterns within strings. On the left, the Unconstrained-Model adheres to logic and a succinct approach. On the right, the Constrained-Model commits logical errors, beginning with the unnecessary concatenation of string $a$ with itself. This discrepancy highlights the decrease in performance that may arise when a model is constrained to use replicas of common functions. The upper part of the figure shows the docstring, which outlines the specifications of the \texttt{cycpattern\_check} function.}
\label{fig:const_failure_example}
\end{figure}

\vspace{4cm}
\newpage
\subsection{Analyzing Failure Cases on $\mathbf{HE_{BP,R}}$} \label{fails}
\begin{table}[!h]
\centering
\small
\caption{Detailed report of failures when prompted to use custom sub-functions.}
\begin{tabularx}{\linewidth}{|c|X|c|}
\hline
\textbf{Task ID} & \textbf{Desc. of Failure} & \textbf{Failure Categorization} \\
\hline
144 & casts a string fraction to a float instead of splitting numerator and denominator first & logic \\
\hline
65 & does not reverse string if shift \% length = 0 (should always reverse when shift $>$ len) & logic \\
\hline
124 & assumes string formatting is correct (can be split on the ‘-’) ; does not account for leap years & logic \\
\hline
149 & sorts by string value rather than string length and alphabetically & logic \\
\hline
114 & restarts sub-array if element is negative rather than if it is less than current element &  logic \\
\hline
79 & does not remove the 0b prefix from the convert\_to\_binary func. & logic (didn’t understand custom func well) \\
\hline
142 & does not include the case where index is not 3 or 4 & logic \\
\hline
121 & selects odd position elements instead of even by starting at index 1 instead of 0 & logic \\
\hline
89 & does not mod by 26 to ensure staying within alphabet & logic \\
\hline
97 & needs to separate modulo operations by a parentheses & logic \\
\hline
154 & does not correctly check rotations of b & logic \\
\hline
99 & always rounds negative numbers toward 0, even when decimal part is greater than or equal to -0.5 & logic \\
\hline
110 & returns "yes" prematurely after checking if lst2 has even elements & logic \\
\hline
84 & converts char to bin then bin to int instead of char to int & logic \\
\hline
70 & does not recompute min and max value within the while loop & logic \\
\hline
82 & used get\_ceiling() when floor should’ve been used & logic \\
\hline
90 & does not account for multiple smallest elements (i.e. [1,1] case) originally avoided by converting list to a set & logic \\
\hline
\end{tabularx}
\label{fig:failures}

\end{table}

\newpage
\subsection{Determining the Optimal Number of Sub-Functions to Provide, Per Question}  \label{ablations}
\begin{figure}[!h]
    \centering
    \includegraphics[width = 1\textwidth]{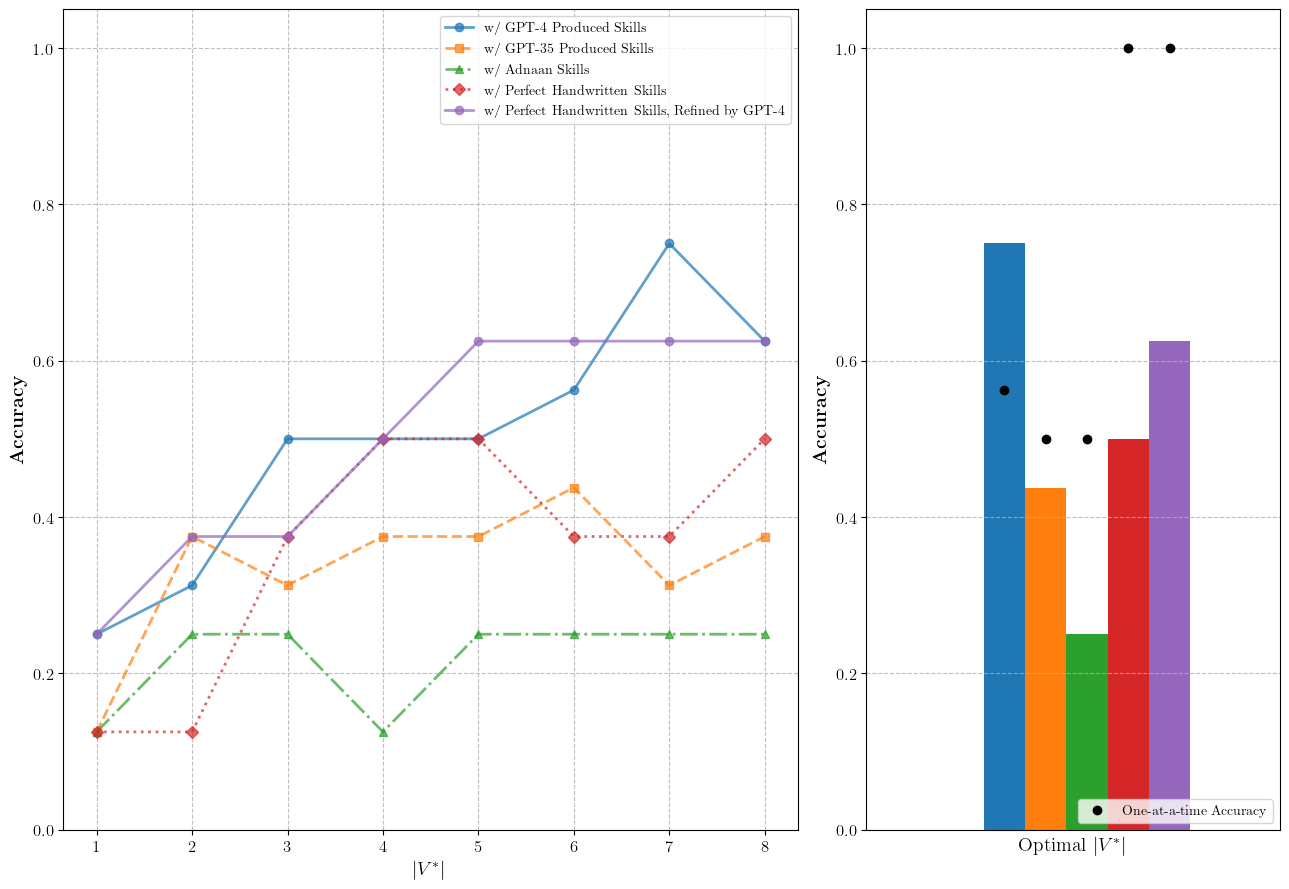}
    \caption{\textbf{Impact of $\mathbf{|V^*|}$ on Performance in $\mathbf{HE_{CF}}$}. This figure analyzes the relationship between the number of functions included in $V^*$ and performance on the $HE_{CF}$ dataset, using functions proposed by GPT-4 (blue), GPT-3.5 (orange), and a human expert (red). The batch of functions marked in purple are the perfect-handwritten skills after being refined by GPT-4. Results for sub-skills proposed by a non-expert human (Adnaan) are presented in green. The black dot on the right bar plot represents the naive expected accuracy when the optimal number of sub-skills are included, which should (naively) match the average accuracy across all 8 questions when only a single function is included in the constrained model's context when solving the related question. The model is only able to achieve 50\% of its one-at-a-time accuracy when all human-written functions are provided (red bar plot). But when providing all functions generated by GPT-4 the model outperforms its one-at-a-time accuracy (blue bar plot).}
    \label{fig:ablations}
\end{figure}
In our experiments, for a given question only the sub-function generated with respect to the question is provided to the constrained model on its subsequent attempt at solving the given question (i.e. \(|V^*| = 1\)). Here, we explore how varying the number of functions in \(V^*\) influences model performance. We consider four-sets of 8 functions generated for each \(HE_{CF}\) question by GPT-4, GPT-3.5, a non-expert human (Adnaan), and a human-expert. In each set (e.g., GPT-4's proposed set) of 8 proposed sub-functions, each sub-function is proposed with the intention of supporting only one unique problem in \(HE_{CF}\). For each set of 8 proposed functions, we vary the inclusion of these functions (from 1 to 8) within \(V^*\) and evaluate replica-constrained GPT-3.5 performance on all questions in \(HE_{CF}\). When a sub-function is ablated from \(V^*\), it is also made absent from \(V\), meaning that the model is not provided the sub-function in any way. Performance is compared against the average accuracy across all 8 questions obtained when providing a sub-function solely to its corresponding \(HE_{CF}\) question. We call this the 'one-at-a-time-accuracy'. The naive assumption is that when all functions are integrated into \(V^*\), the accuracy should mirror the average 'one-at-a-time' configuration. 

Our results, presented in Figure \ref{fig:ablations} show that this is not the case. We find that when the optimal amount of functions included (i.e., the $|V^*|$ which maximises accuracy), there is variability in how the model performs in comparison to the expected one-at-a-time accuracy. For example, when using the ``perfect" set of handwritten functions (deemed perfect because when the functions are fed one-at-a-time to their intended question, all questions pass) the model only achieves 50\% of its one-at-a-time accuracy when all functions are provided (red bar plot). But when providing all functions generated by GPT-4, the model achieves 120\% of its one-at-a-time accuracy (blue bar). In general, we see that GPT-generated functions, when provided to a constrained model alongside other functions, result in better performance than when handwritten functions are fed alongside other functions. Due to the small sample size of questions (8), this can be attributed to the stochastic nature of GPT-3.5 (i.e., arbitrary variations in the prompt produce better/worse performance). However, if this is not the case, then this implies that the GPT models are formulating functions in a way that, when fed among a batch of other functions, allows the LM to better use the right function for the right problem. 

To test this hypothesis, we took all functions in the set of expert-written functions (red) and asked GPT-4 to "Make this function more understandable." We then evaluated the results of this new set of refined functions, shown in purple. Although the refined set results in better performance than the non-refined set, the refined set only achieves 62\% of its one-at-a-time accuracy. This results indicates that the functions in the handwritten set overfit the one-at-a-time scenario where they are the only sub-function being provided to the model. This means that they are perfectly crafted to trigger the model to output the correct solution when fed alone, but when combined with other functions (i.e., a variation in the prompt) the correct response isn't generated, implying that the human-written functions do not generalise as well as GPT-generated functions to scenarios where multiple functions are provided at once.

\newpage
\subsection{Sub-Datasets} \label{datasets}
\begin{figure}[!h]
    \centering
    \includegraphics[width=\textwidth]{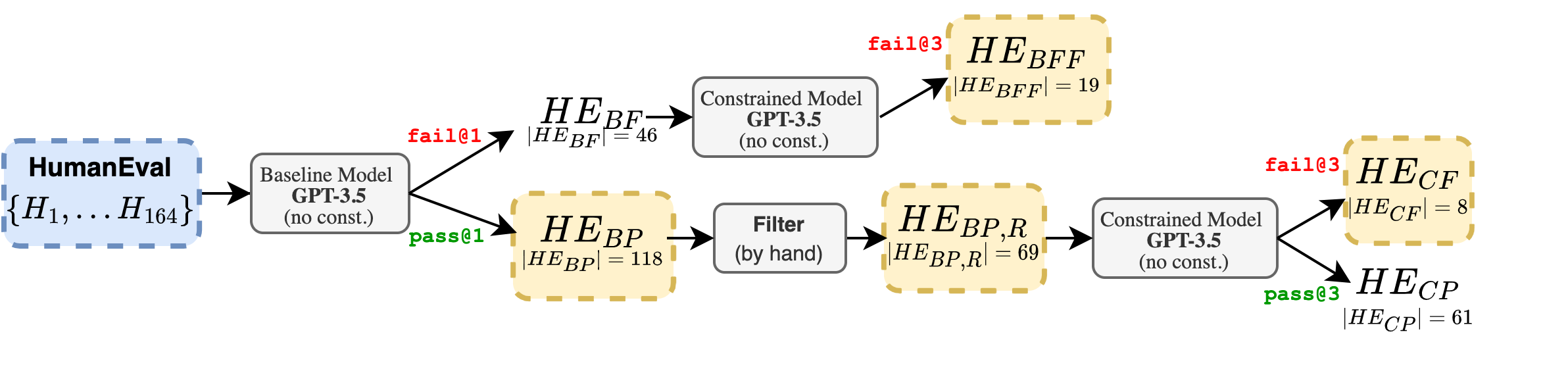}
    \caption{\textbf{Sub-datasets extracted from HumanEval.} 
Relevant sub-datasets are highlighted in yellow. \(HE_{BFF}\) includes the questions that the original, unconstrained model failed to solve with just one attempt. On the other hand, \(HE_{CF}\) comprises questions that the baseline, unconstrained model answered correctly, but the constrained model could not solve. Both of these sub-datasets undergo a refinement process, where the constrained baseline model is allowed three more attempts at answering all questions (with \(temp = 0.5\)). This step ensures that any future successes in our experiments are more likely a result of genuine improvement rather than random chance.}
    \label{fig:subdatasets}
\end{figure}

Four sub-datasets are extracted from HumanEval and used in our experiments. One sub-dataset, $APPS_{BP}$, is extracted from APPS. Refer to Fig. \ref{fig:subdatasets} for a visual guide explaining the extraction of the HumanEval sub-datasets. 

$\mathbf{HE_{BP}.}$  The first sub-dataset, \(HE_{BP}\), consists of 118 questions. These questions are sourced from those that the baseline model, GPT-3.5, can answer in one attempt using the "half-shot" pass@1 method, without any constraints. The aim of \(HE_{BP}\) is to pinpoint questions correctly addressed by the original unconstrained model.

$\mathbf{HE_{BP,R}.}$ The second sub-dataset, \(HE_{BP,R}\), comprises 69 questions. These are extracted from \(HE_{BP}\) by manually identifying questions whose answers contain at least one free-standing function call, such as \texttt{len(.)} or \texttt{math.sqrt(.)}. The function calls identified serve as a reference to construct \(V_{rep}\), as detailed in Table \ref{tab:replicas}. The primary goal of \(HE_{BP,R}\) is to spot questions from the unconstrained model that feature a free-standing function call in their solutions. This set helps formulate \(V_{rep}\).

$\mathbf{HE_{CF}.}$ The third sub-dataset, \(HE_{CF}\), encompasses 8 questions. These are derived by testing the baseline GPT-3.5, but this time constrained on \(V_{rep}\), on \(HE_{BP,R}\) with a three-attempt limit, and collecting all the unanswered queries. \(HE_{CF}\) is designed to ensure that if our method improves a model's performance on a question from this set during tests, it is likely due to our specific improvements and not mere randomness. This dataset aids in gauging the effectiveness of introducing sub-functions.

$\mathbf{HE_{BFF}.}$ The fourth sub-dataset, \(HE_{BFF}\), has 19 questions, obtained in a two-step manner:

\begin{enumerate}
    \item Initially, we gather questions the baseline model, GPT-3.5, fails to answer in one attempt using the "half-shot" pass@1 method, without constraints.
    \item Following that, we test the constrained GPT-3.5 on \(V_{rep}\) on the previously collected questions, allowing for three tries, and select all the incorrectly answered ones.
\end{enumerate}

The purpose of $HE_{BFF}$ is to represent the questions that the original, unconstrained model couldn't solve in one attempt (as described in step 1). In our experiments, we want to decrease the likelihood that the questions in this dataset are passed just by random chance, rather than because of the improvements introduced by our method. To do this, we use the process in step 2 to refine and confirm the questions that were selected from step 1.

\textbf{$\mathbf{APPS_{BP}}.$}  $APPS_{BP}$ consists of 1329 questions from the APPS dataset which the unconstrained GPT-3.5 does not pass with one attempt (i.e., using ``half-shot" pass@1). This sub-dataset is similar to $HE_{BP}$, but derived from APPS rather than HumanEval. 
\end{document}